\pdfoutput=1

\documentclass[11pt]{article}

\usepackage[final]{acl}
\usepackage{hyperref}

\usepackage{times}
\usepackage{latexsym}

\usepackage[T1]{fontenc}

\usepackage[utf8]{inputenc}

\usepackage{microtype}
\usepackage[normalem]{ulem}
\usepackage{inconsolata}
\usepackage{xcolor}
\definecolor{fullgreen}{rgb}{0.502, 0.788, 0.643}
\definecolor{fullred}{rgb}{0.800, 0.447, 0.541}
\usepackage{graphicx}

\usepackage{enumitem}

\usepackage{multirow}
\usepackage{amssymb}
\usepackage{color}
\usepackage{amsmath}
\usepackage{mdframed}
\usepackage{bm}
\usepackage{arydshln}
\usepackage{tcolorbox}
\usepackage{algorithm}
\usepackage{algorithmicx}
\usepackage[noend]{algpseudocode}
\usepackage{subcaption}
\usepackage[export]{adjustbox}
\usepackage{rotating}
\usepackage{booktabs}
\usepackage{marvosym}
\usepackage{footmisc}
\usepackage{colortbl}
\usepackage{pifont}
\usepackage{svg}
\newcommand{\shline}{\hline\hline}

\definecolor{mygray}{RGB}{240,240,240}
\definecolor{ForestGreen}{RGB}{34,139,34}
\definecolor{twiggreen}{RGB}{221,238,214}
\newcommand{\fg}[1]{\mathbf{\mathcolor{ForestGreen}{#1}}}
\definecolor{Forestred}{RGB}{220,50,50}

\title{CROP: Contextual Region-Oriented Visual Token Pruning}

\author{First Author \\
  Affiliation / Address line 1 \\
  Affiliation / Address line 2 \\
  Affiliation / Address line 3 \\
  \texttt{email@domain} \\\And
  Second Author \\
  Affiliation / Address line 1 \\
  Affiliation / Address line 2 \\
  Affiliation / Address line 3 \\
  \texttt{email@domain} \\}
  
 \author{Jiawei Guo\textsuperscript{1,2}, Feifei Zhai\textsuperscript{\Letter 1}, Pu Jian\textsuperscript{1,2}, Qianrun Wei\textsuperscript{1,2}, Yu Zhou\textsuperscript{1,3} \\
  $^1$ State Key Laboratory of Multimodal Artificial Intelligence Systems, \\Institute of Automation, CAS, Beijing, China \\
  $^2$ School of Artificial Intelligence, University of Chinese Academy of Sciences, Beijing, China \\
  $^3$ Fanyu AI Laboratory, Zhongke Fanyu Technology Co., Ltd, Beijing, China \\
  {\texttt \{guojiawei2024, feifei.zhai, jianpu2023, weiqianrun2025\}@ia.ac.cn}, \texttt yzhou@nlpr.ia.ac.cn
  }

\begin{document}
\maketitle

\DefineFNsymbols*{1}{\Letter}
\setfnsymbol{1}

\renewcommand{\thefootnote}{\fnsymbol{footnote}} 
    \footnotetext[1]{Corresponding Author}
\renewcommand{\thefootnote}{\arabic{footnote}}

\begin{abstract}
Current VLM-based VQA methods often process entire images, leading to excessive visual tokens that include redundant information irrelevant to the posed question. This abundance of unnecessary image details creates numerous visual tokens, drastically increasing memory and computational requirements in VLMs. To address this, we propose Contextual Region-Oriented Visual Token Pruning (CROP\footnote{The related code and dataset are released at: \url{https://github.com/JiaweiGuo98/CROP}}), a novel framework to compress visual tokens through a two-step process: Localization and Pruning. Specifically, CROP first employs an efficient model to identify the contextual region relevant to the input query. Subsequently, two distinct strategies are introduced for pruning: (1) Pre-LLM Compression (PLC), which adaptively compresses different image regions with varying ratios, and (2) Inner-LLM Pruning (ILP), a training-free method that prunes tokens within early LLM layers guided by the identified contextual region. Extensive experiments on a wide range of VQA tasks demonstrate that CROP significantly outperforms existing visual token pruning methods and achieves state-of-the-art performance.

\end{abstract}

\section{Introduction}

Recently, visual question answering (VQA) have achieved remarkable progress due to the rapid development of vision language Models (VLMs) \citep{bai2023qwentechnicalreport, yin2023survey, li2024minigeminiminingpotentialmultimodality, ren2024timechattimesensitivemultimodallarge, DBLP:conf/cvpr/SinghNSJCBPR19, zhou2025mlvubenchmarkingmultitasklong}. Current VLM-based VQA methods utilize information from the entire image, but for specific questions, we need to locate local image regions to support the answer. Moreover, redundant image information also introduce a large number of visual tokens, requiring much higher memory and computation in VLMs \citep{DBLP:journals/corr/abs-2501-03895, DBLP:journals/corr/abs-2412-00876}. For example, there are 576 visual tokens in LLaVA-1.5 \citep{liu2023visualinstructiontuning}, and the number is 2880 for a 672*672 image in LLaVA-NeXT \citep{yang2024visionziplongerbetternecessary}.

\begin{figure*}[t]
    \centering
    \includegraphics[width=0.98\linewidth]{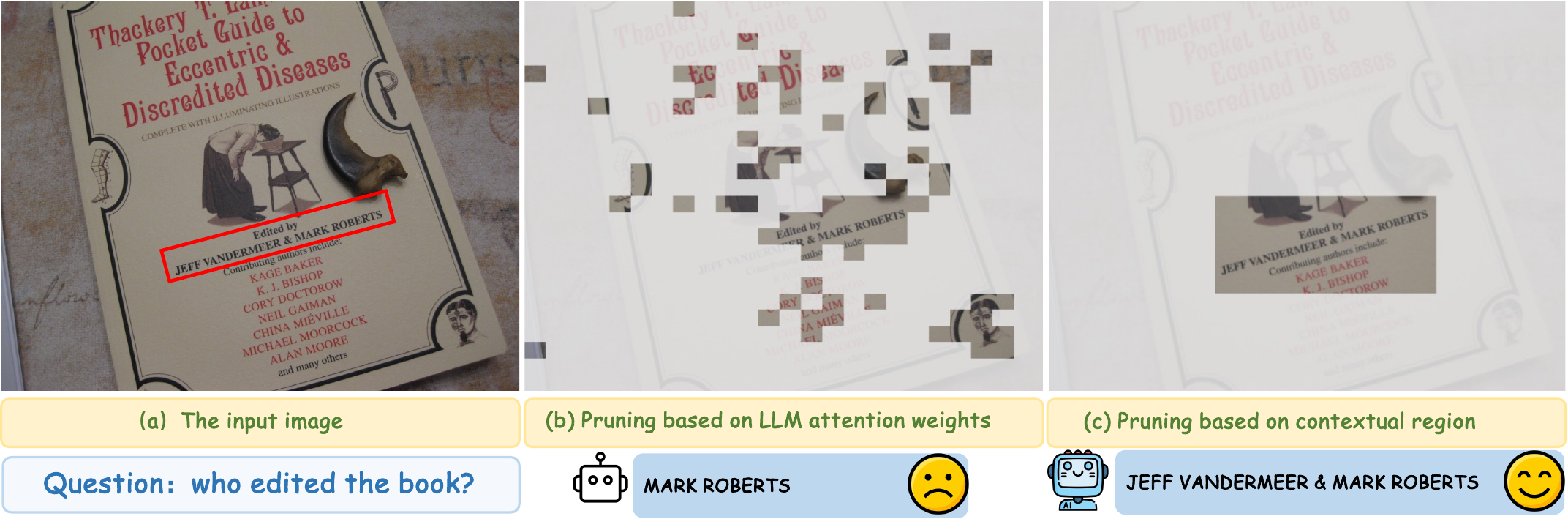}
    \caption{Comparison of different visual token pruning. (a) The input image and question. The red rectangle shows the region containing the answer of the question. (b) The visualization of retained tokens by attention-based pruning method. (c) The visualization of identified contextual region by our CROP framework.}
    \label{figure: Comparison}
    \vspace{-2mm}
\end{figure*}

For example in Figure~\ref{figure: Comparison}(a), we actually only need a small region to answer the question, but still have to transform the entire image into so many visual tokens. To overcome this problem, visual token pruning methods have been emerged. Some methods reduce visual tokens before inputting them into LLM \citep{DBLP:journals/corr/abs-2412-00876}, which primarily depend on intrinsic image semantics. Others perform pruning inside the early LLM layers, usually on the basis of attention map \citep{DBLP:journals/corr/abs-2403-06764}. However, these methods fail to account for the input question, which might ignore the key task-relevant information \citep{li2024tokenpackerefficientvisualprojector}.

In this paper, we introduce a novel framework, Contextual Region-Oriented Vision Token Pruning (CROP), to facilitate effective visual token pruning. We define the contextual region as a contiguous visual area of the input image that captures the key information for answering the question. The proposed CROP framework comprises two stages: Localization and Pruning. During the Localization Stage, an efficient model identifies the contextual region. Subsequently, in the Pruning Stage, this identified region serves as a key information source to guide two simple yet effective pruning strategies: Pre-LLM Compression (PLC) and Inner-LLM Pruning (ILP).

In PLC method, we propose a compression module that adaptively adjusts the compression ratio, assigning a lower compression ratio to the contextual region obtained from the localization stage, and a higher ratio otherwise. The ILP method prunes visual tokens in the LLM layers by the guidance of L-CR. Extensive experiments on a wide range of VQA benchmarks show that the proposed PLC and ILP methods consistently outperform the existing compression techniques, and achieve the state-of-the-art performance.

The main contributions of this paper are summarized as follows:
\vspace{-0.5em}
\begin{itemize}
\setlength{\itemsep}{0pt}
\setlength{\parsep}{0pt}
\setlength{\parskip}{0pt}
\item We develop an efficient Localization model to identify contextual regions, which serve as a plug-and-play component for guiding visual token pruning.
\item We introduce a novel contextual region-oriented vision token pruning framework, and develop two different approaches on pruning, PLC and ILP. To our best knowledge, this is the first work of introducing contextual regions for visual tokens pruning.
\item Experimental results demonstrate that our proposed method achieves state-of-the-art performance without requiring any training or fine-tuning in ILP method.
\end{itemize}
\vspace{-0.5em}

\section{Related Works}

\subsection{Large Language Models}

The advancement of Large Language Models (LLMs) has redefined state-of-the-art performance across a vast landscape of tasks, spanning foundational natural language understanding \citep{jing2024dq, zhang2025discoveringsemanticsubdimensionsdisentangled, chen2025ladmlongcontexttrainingdata, chen2025lr2benchevaluatinglongchainreflective}, machine translation \citep{guan-etal-2025-trifine, liang2024document, zhang2025understand}, and complex reasoning in domains like mathematics and agent-based systems \citep{sun2025ktae, xu2025hit}. A particularly impactful frontier has been their expansion into multimodality, enabling breakthroughs in audio processing \citep{diao2025learning, diao-etal-2024-learning, diao2025soundmind} and, crucially for this work, visual understanding \citep{jian2025lookagainthinkslowly, jian2025teaching, jian2024large}, which has given rise to powerful VLMs. While these VLMs excel at many tasks, their success comes at a high computational cost due to the large number of visual tokens processed from each image, motivating the need for efficient token reduction strategies.

\subsection{Vision Token Reduction in VLMs}

In recent years, a substantial amount of research has emerged focusing on visual token pruning and compression, with these methods emphasizing the use of attention mechanisms within VLMs to retain important visual information \citep{han2025filtercorrelatecompresstrainingfree, ye2025vocollamavisioncompressionlarge, yan2024tgllavatextguidedllava, DBLP:journals/corr/abs-2410-17247, DBLP:journals/corr/abs-2411-10803}. For instance, FastV \citep{DBLP:journals/corr/abs-2403-06764} leverages cross-modal attention from intermediate layers of the model to preserve the Top-$R$ visual tokens, while TwigVLM \citep{shao2025growingtwigacceleratelarge} introduces additional trainable layers to improve the precision of token selection. VisionZip \citep{yang2024visionziplongerbetternecessary} aims to use visual encoder attention to retain, compress, and merge key visual information. SparseVLM \citep{DBLP:journals/corr/abs-2410-04417} employs bidirectional selection of both text and visual tokens to extract more informative visual representations. LLaVA-Mini \citep{DBLP:journals/corr/abs-2501-03895} incorporates an additional fusion module between the encoder and the LLM to perform early cross-modal fusion. However, these methods often result in discrete and fragmented visual tokens, disrupting the spatial semantics and continuity of visual regions. In contrast, our CROP method preserves continuous contextual regions directly relevant to the input question

\begin{figure*}[t]
    \centering
    \includegraphics[width=0.98\linewidth]{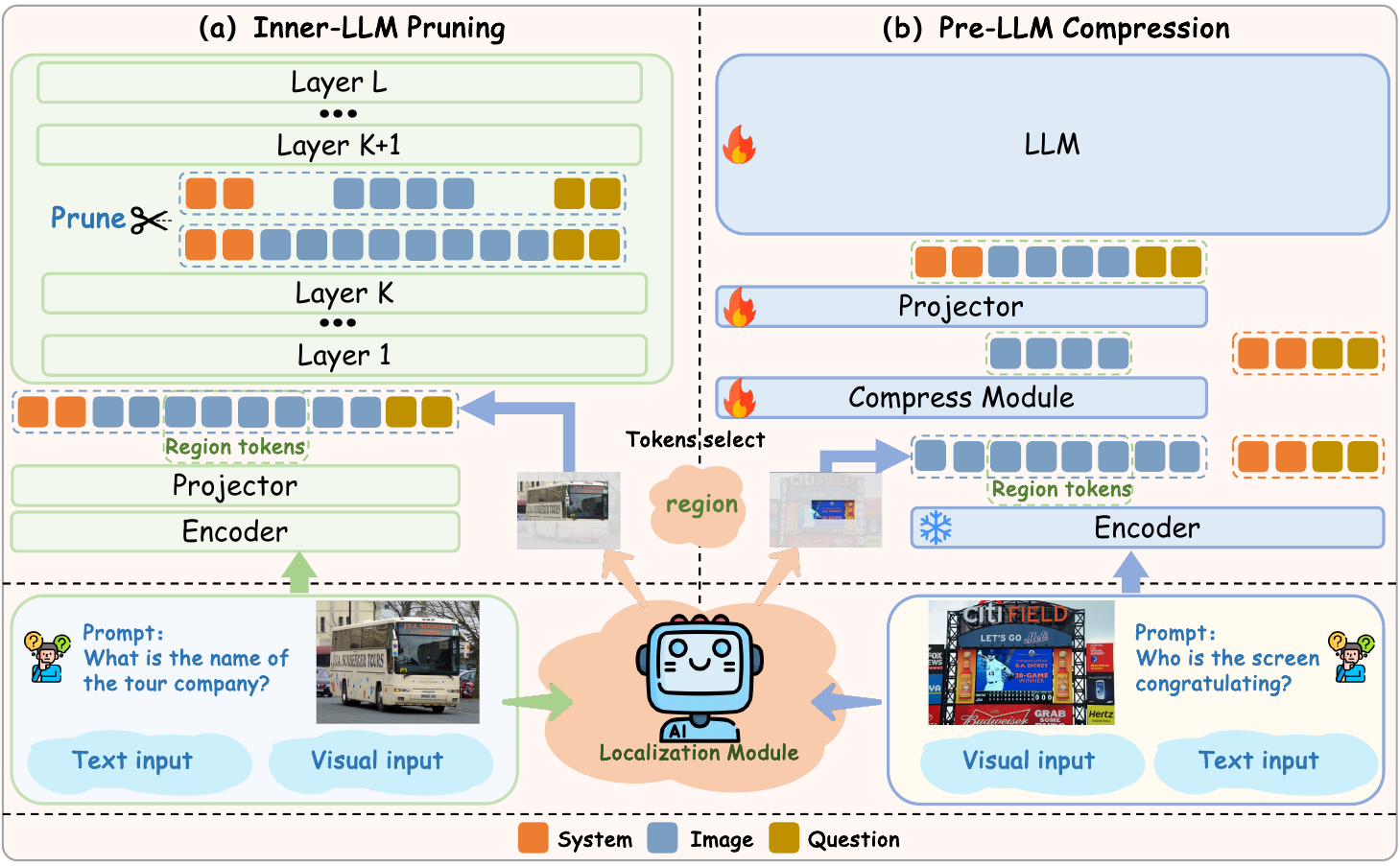}
    \caption{Overview of the CROP framework. Given achen2025lr2benchevaluatinglongchainreflective visual input and a textual prompt, the Localization Module first identifies task-relevant contextual regions. This regional information then guides two distinct visual token compression strategies: (a) ILP, where tokens outside the contextual region are pruned directly within an early layer (Layer K) of the main VLM without retraining it; and (b) PLC, where a dedicated compression module processes tokens from contextual and non-contextual regions before they are fed to the LLM. The visual input pipeline and the LLM are shown in context for both strategies.}
    \label{figure: Architecture}
    \vspace{-5mm}
\end{figure*}

\subsection{Contextual Region Perception in VLMs}

Recent advancements have shown that VLMs possess strong capabilities in identifying and grounding information within specific visual regions. For example, \citet{zhang2025mllmsknowlooktrainingfree} demonstrated that VLMs often know which visual areas to focus on, even when answering questions incorrectly. \citet{shao2024visualcotadvancingmultimodal} found that cropping and re-inputting relevant regions enhances visual perception, introducing datasets for localization. Additionally, VPT \citep{yu2025introducingvisualperceptiontoken} replaced precise coordinates with region selection tokens, improving localization accuracy. These studies suggest that VLMs have significant abilities in question-guided region localization \citep{li2025migicianrevealingmagicfreeform}. Our work builds on this by fine-tuning VLMs for explicit region identification, guiding visual token pruning and compression to preserve the most contextually relevant visual information.

\begin{figure}[t]
    \centering
    \includegraphics[width=\columnwidth]{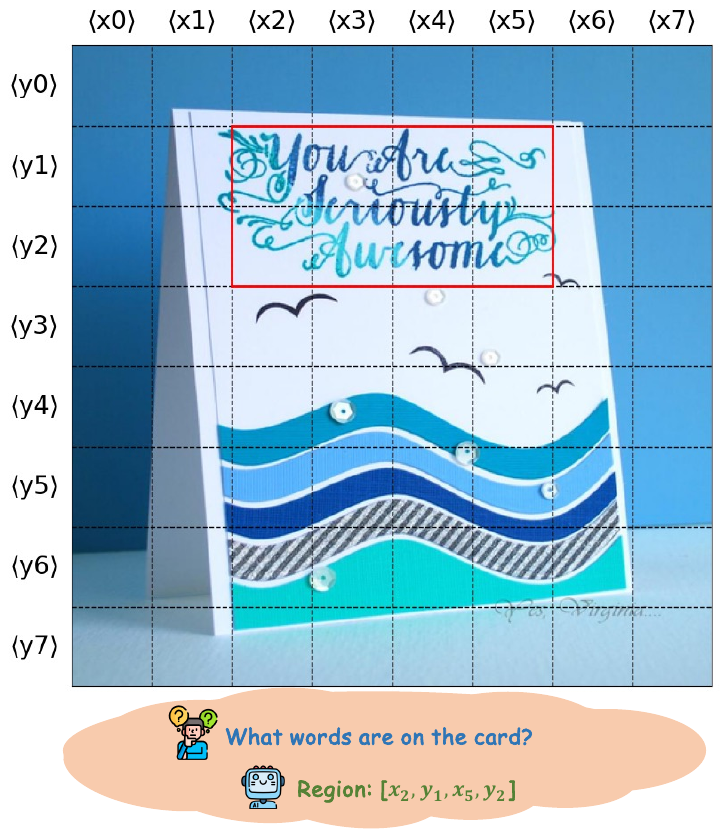}
    \caption{Illustration of the 8×8 Grid-based Region Definition for Contextual Region Localization. The input image is overlaid with an 8×8 grid. For a given question (e.g., "What words are on the card?"), the Localization Module identifies the contextual region containing the answer (e.g., the text "You Are Seriously Awesome"). This region is represented by the minimum and maximum 0-indexed block coordinates $\left[x_{min}, y_{min}, x_{max}, y_{max}\right]$, which in this example corresponds to $\left[x_{2}, y_{1}, x_{5}, y_{2}\right]$.}
    \label{figure: Coords}
    \vspace{-5mm}
\end{figure}

\subsection{Chain-of-thought Reasoning}

Chain-of-Thought \citep{wei2023chainofthoughtpromptingelicitsreasoning} (CoT) prompting has demonstrated that intermediate reasoning steps improve problem-solving in large language models. While this approach has been extended to the vision domain, VLMs still show limited ability to effectively localize and interpret visual regions. To overcome this limitation, recent methods have introduced locate-then-answer paradigms. For example, \citet{wu2023vguidedvisualsearch} fine-tunes VLMs to use a visual search model when additional localization is necessary, and \citet{luan2024textcotzoomenhancedmultimodal} mimics human scanning by first generating a global description before localizing regions. Other approaches, such as \citet{man2025argusvisioncentricreasoninggrounded, ge2025mrfdmultiregionfusiondecoding}, employ Mixture-of-Experts image encoder or cross-attention across VLMs to refine predictions. However, these methods often rely on discrete attention mechanisms or precise bounding box regression, which can struggle with sustained focus and require costly model-specific fine-tuning. In contrast, our approach uses a lightweight VLM for localization that predicts coordinates at a coarse, block-level granularity. This strategy not only achieves more robust localization but also allows our module to be integrated into various VLMs without requiring any retraining.

\section{Method}
\subsection{Overview}

To preserve the completeness of contextual regional information during visual token processing—a factor we hypothesize is crucial for robust understanding compared to methods that might select sparse, disconnected tokens—we introduce CROP. The overall architecture of CROP is illustrated in Figure~\ref{figure: Architecture}. The framework operates in two main stages: first, a dedicated Localization Module identifies contextual regions pertinent to a given query. Second, this regional information guides the subsequent visual token processing in the main VLMs. This approach emphasizes retaining comprehensive visual information within these contextual regions, particularly their inherent spatial structures and the relationships between constituent patches.

The main VLMs used in our experiments typically process an input image into a grid of visual tokens (e.g., 24×24=576 tokens from ViT-L/14 \citep{DBLP:conf/icml/RadfordKHRGASAM21}), which maintain spatial correspondence with the original image patches. This spatial mapping is crucial as it allows CROP to accurately translate identified contextual regions into specific sets of visual tokens for targeted processing. Losing this explicit mapping, as might occur in some global pooling or aggressive token merging strategies, could obscure fine-grained spatial details vital for many VQA tasks. We instantiate and evaluate CROP through two distinct strategies: ILP and PLC. ILP directly removes visual tokens from non-contextual regions within the early layers of the main VLMs, requiring no retraining. PLC employs a lightweight module to compress visual tokens before they enter the LLM, prioritizing the fidelity of contextual regional information while integrating broader context.

\subsection{Contextual Region Localization}

A lightweight VLM serves as an independent Localization Module, which pre-identifies contextual regions to provide guidance to the main VLM. Direct bounding box generation by VLMs can be challenging, especially across varied image resolutions. However, guiding models to identify broader regional extents can yield more robust localization. Inspired by this and approaches like Visual Prompt Tuning \citep{yu2025introducingvisualperceptiontoken} and Visual-cot \citep{shao2024visualcotadvancingmultimodal}, we developed a specialized training dataset to enhance the Localization Module's ability to discern salient information based on spatial layout.

\begin{table*}[t]
    \centering
    \resizebox{\linewidth}{!}{%
    \begin{tabular}{@{}l l@{}}
    \toprule
    \textbf{Function} & \textbf{Dataset} \\ 
    \midrule
    \multirow{3}{*}{Region Localization (410k)} 
        & TextVQA (73k), Flickr30k (136k), DocVQA (33k) \\
        & VSR (3k), GQA (88k), OpenImage (43k) \\
        & CUB (4k), V7W (30k) \\
    \midrule
    Preserving Instruction Following Ability (200k) 
        & LLaVA Instruction Tuning data(200k) \\
    \bottomrule
    \end{tabular}%
    }
    \caption{Composition of the training dataset. Our training dataset includes the Region Localization samples used to train the Localization and Compress Modules, as well as the LLaVA Instruction Tuning data aimed at preserving the model's original instruction-following capability. In total, the training dataset comprises 610k samples.}
    \label{table: datasets}
\end{table*}

We represent images on an 8×8 grid, as depicted in Figure~\ref{figure: Coords}. A rectangular contextual region within this grid is defined by block coordinates $\left[x_{min}, y_{min}, x_{max}, y_{max}\right]$. These are 0-indexed, meaning each coordinate can range from 0 to 7 inclusive, and must satisfy $0 \le x_{min} \le x_{max} \le 7$ and $0 \le y_{min} \le y_{max} \le 7$.

We selected the Qwen2VL-2B model \citep{wang2024qwen2vlenhancingvisionlanguagemodels} as our Localization Module and fine-tuned it on this curated dataset. This module, requiring approximately 8 hours of training, achieves high localization accuracy and can be seamlessly integrated as a plug-and-play component with various large VLMs.

\subsection{Visual Token Compression}

Given the localized contextual regions, CROP implements ILP and PLC. Our experiments primarily utilize the LLaVA model family \citep{liu2023visualinstructiontuning, liu2024improvedbaselinesvisualinstruction}, which employs a ViT \citep{DBLP:conf/iclr/DosovitskiyB0WZ21} for visual encoding. In such models, an input image $I^{v}$ is processed by the visual encoder, and its output features are transformed by a projector (e.g., an MLP) into $N^{2}$ visual tokens, $X^{v} \in \mathbb{R}^{N^2 \times d_h}$. These visual tokens are projected into a space compatible with the text prompt tokens $X^{T}$, where $d_h$ is the LLM's hidden embedding dimension. The combined multimodal input for the LLM often follows the structure: 
\begin{equation}
\small
    \left<X^{q}_{1},\dots, X^{q}_{m},X^{v}_{1},\dots, X^{v}_{N^{2}}, X^{q}_{m+1},\dots,X^{q}_{N_{t}} \right> \label{eq:tokens}
\end{equation}
where $\langle X^{q}_1, \dots, X^{q}_m \rangle$ is the system prompt, and $\langle X^{q}_{m+1}, \dots, X^{q}_{N_t} \rangle$ is the user query.

\paragraph{Inner-LLM Pruning.} While some existing methods for reducing visual tokens in VLMs utilize cross-modal attention or introduce additional trainable modules for token selection, our ILP offers a distinct approach. Our observations, consistent with existing literature \citep{DBLP:journals/corr/abs-2403-06764, chen2023llavainteractiveallinonedemoimage}, suggest that VLMs rely more heavily on explicit visual tokens in their earlier layers, with information becoming progressively more abstract and multimodally fused in deeper layers. 

ILP leverages this by using the precise, task-relevant region information from the Localization Module. Based on the identified contextual regions, visual tokens falling outside this area are removed at a designated early layer $K$ of the VLM, as illustrated in Figure~\ref{figure: Architecture}(a). This approach is plug-and-play, as it requires no retraining of the large VLM. Our experiments validate that this early-stage, region-guided pruning achieves state-of-the-art performance on multiple VQA benchmarks, even on VLMs not specifically adapted for CROP.

\paragraph{Pre-LLM Compression.} We hypothesize that contextual regions contain the most critical visual information for query resolution. To explore this while retaining some global context, PLC is designed as an external, trainable compression module that prioritizes contextual region fidelity, depicted in Figure~\ref{figure: Architecture}(b).
 
Visual tokens from the encoder, $X^{v}$, are first partitioned into contextual region tokens $X^{kv}$ and non-contextual region tokens $X^{nkv}$, guided by the Localization Module. We introduce two sets of learnable queries, $\mathit{Q}^{k}$ and $\mathit{Q}^{nk}$, to compress these token sets respectively. Positional encodings $ P(\cdot ) $ are added to queries and tokens before attention. Compression is achieved via scaled dot-product attention (Equations \ref{eq:plc_kv_attn} and \ref{eq:plc_nkv_attn}):
\begin{equation}
        \hat{X}^{kv} = \text{softmax}\left(\frac{P(Q^{k}) (P(X^{kv}))^T}{\sqrt{d_h}}\right) P(X^{kv})
    \label{eq:plc_kv_attn}
\end{equation}
\vspace{-10pt}
\begin{equation}
         \hat{X}^{nkv} = \text{softmax}\left(\frac{P(Q^{nk}) (P(X^{nkv}))^T}{\sqrt{d_h}}\right) P(X^{nkv})
    \label{eq:plc_nkv_attn}
\end{equation}
Here, $\hat{X}^{kv} $ and $\hat{X}^{nkv}$ are the compressed representations.

In addition to the compressed representations, we define anchor tokens $X^{r}$ to preserve fine-grained details. $X^{r}$ is an 8x8 square patch of 64 tokens, extracted from the original visual encoder output, centered around the geometric midpoint of the contextual region $X^{kv}$. These query the compressed contextual region tokens $\hat{X}^{kv}$ for contextual integration, followed by a residual connection:
\begin{equation}
\footnotesize
    X^{fused} = \text{softmax}\left(\frac{P(X^{r}) (P(\hat{X}^{kv}))^T}{\sqrt{d_h}}\right) P(\hat{X}^{kv}) + X^{r}
\label{eq:plc_fuse_attn}
\end{equation}
The final compressed visual representation $\hat{X}^{v} $ concatenates the fused contextual region tokens with the compressed non-contextual region tokens:
\begin{equation}
        \hat{X}^{v} = \text{Concat}(X^{fused}, \hat{X}^{nkv})
    \label{eq:plc_final_concat}
\end{equation}
This PLC strategy aims to balance substantial token reduction with the  preservation of essential regional details and broader contextual cues. 

\section{Experiments}

\begin{table*}[t]
    \centering
    \setlength{\tabcolsep}{3.5pt}
    \renewcommand{\arraystretch}{1.3}
    \small
    \centering
    \vspace{2mm}
    \resizebox{1\textwidth}{!}{
    \begin{tabular}{l|cccccc|c|cccccc|c}
        \shline
        & \multicolumn{7}{c|}{\textbf{LLaVA-1.5-7B}} & \multicolumn{7}{c}{\textbf{LLaVA-1.5-13B}} \\
        \cline{2-15}
        \textbf{Method} & 
        \footnotesize{\textbf{GQA}} &
        \footnotesize{\textbf{MME}} & 
        \footnotesize{\textbf{VQA}$^{\text{T}}$} &
        \footnotesize{\textbf{SQA}} &   \footnotesize{\textbf{VQA}$^{\text{V2}}$} & \footnotesize{\textbf{POPE}} & \footnotesize{\textbf{Avg.}} & \footnotesize{\textbf{GQA}} &
        \footnotesize{\textbf{MME}} & 
        \footnotesize{\textbf{VQA}$^{\text{T}}$} &
        \footnotesize{\textbf{SQA}} &   \footnotesize{\textbf{VQA}$^{\text{V2}}$} & \footnotesize{\textbf{POPE}} & \footnotesize{\textbf{Avg.}} \\
        \shline
        \rowcolor{mygray}
        \multicolumn{15}{c}{\footnotesize{\textit{Upper Bound, 576 Tokens} \ (\textbf{100\%})}} \\
        LLaVA-1.5 & 61.9 & 1862 & 58.2 &  69.5 & 78.5 & 85.9 & 100\% & 63.2 & 1818 & 61.3 & 72.8 & 80.0 & 85.9 & 100\% \\
        \hline
        \rowcolor{mygray}
        \multicolumn{15}{c}{\footnotesize{\textit{Retain Averaged 192 Tokens} \ $\fg{(\downarrow 66.7\%)}$}} \\
        \footnotesize{FastV} & 56.5 & 1786 & 57.3 & \textbf{69.5} & 74.6 & 79.2 & 95.5\% & 60.3 & \underline{1807} & \underline{60.4} & \underline{74.0} & 77.7 & 82.3 & 98.0\% \\
        \footnotesize{SparseVLM} & 57.6 & 1721 & 56.1 & 69.1 & 75.6 & 83.6 & 95.8\% & 58.7 & 1768 & 45.4 & 73.1 & - & 82.2 & 92.1\% \\
        \footnotesize{PDrop} & 57.3 & 1797 & 56.5 & 69.2 & 75.1 & 82.3 & 96.2\% & 61.3 & 1663  & \textbf{60.7} & 73.6 & \underline{78.7} & 84.8 & 97.6\% \\
        \footnotesize{VisionZip} & 59.3 & 1783 & 57.3 & 68.9 & 76.8 & \textbf{85.3} & 97.7\% & 59.1 & 1754 & 59.5 & 73.5 & 78.1  & \underline{85.1} & 97.5\% \\
        \footnotesize{VisionZip\ddag} & \underline{60.1} & \textbf{1834} & \textbf{57.8} & 68.2 & \underline{77.4} & \underline{84.9} & \underline{98.4\%} & \underline{61.6} & 1790 & 59.9 & 72.7 & 78.6 & 84.5 & \underline{98.4}\% \\
        \hline
        \textbf{\footnotesize{CROP-ILP}} & \textbf{61.3} & \underline{1817} & \underline{57.4} & \underline{69.2} & \textbf{77.7} & \textbf{85.3} & \textbf{98.9\%} & \textbf{62.7} & \textbf{1822} & \underline{60.4} & \textbf{74.2} & \textbf{78.9} & \textbf{86.2} & \textbf{99.8\%} \\
        \hline
        \rowcolor{mygray}
        \multicolumn{15}{c}{\footnotesize{\textit{Retain Averaged 128 Tokens} \ $\fg{(\downarrow 77.8\%)}$}} \\
        \footnotesize{FastV} & 53.0 & 1646 & 56.0 & \textbf{69.5} & 69.2 & 73.2 & 90.6\% & 57.5 & 1758 & 58 & 73.8 & 74.3 & 79.3 & 94.8\% \\
        \footnotesize{SparseVLM} & 56.0 & 1696 & 54.9 & 67.1 & 73.8 & 80.5 & 93.4\% & 57.9 & \textbf{1774} & 49.9 & 69.9 & - & 81.1 & 92.2\% \\
        \footnotesize{PDrop} & 57.1 & 1761 & 56.6 & 68.4 & 72.9 & 82.3 & 95.2\% & \underline{61.0} & 1490 & \textbf{60.2} & 73.3 & \textbf{78.2} & 83.6 & 95.4\% \\
        \footnotesize{VisionZip} & 57.6 & 1762 & 56.8 & \underline{68.9} & 75.6 & 83.2 & 96.3\% & 57.9 & 1743 & 58.7 & \underline{74.0} & 76.8 & \underline{85.2} & 96.7\% \\
        \footnotesize{VisionZip\ddag} & \underline{58.9} & \textbf{1823} & \textbf{57.0} & 68.3 & \underline{76.6} & \underline{83.7} & \underline{97.4\%} & 60.1 & 1736 & 59.2 & 73.0 & 77.6 & 83.8 & \underline{97.0\%} \\
        \hline
        \textbf{\footnotesize{CROP-ILP}} & \textbf{60.8} & \underline{1771} & \underline{56.9} & \textbf{69.5} & \textbf{76.8} & \textbf{84.4} & \textbf{97.9\%} & \textbf{61.6} & \underline{1768} & \underline{59.4} & \textbf{74.1} & \underline{78.0}  & \textbf{85.8} & \textbf{98.5\%} \\
        \hline
        \rowcolor{mygray}
        \multicolumn{15}{c}{\footnotesize{\textit{Retain Averaged 64 Tokens} \ $\fg{(\downarrow 88.9\%)}$}} \\
        \footnotesize{FastV} & 44.1 & 1218 & 50.7 & 70.0 & 52.0 & 55.6 & 75.9\% & 50.1 & 1408 & 52.2 & 73.2 & 61.1 & 69.3 & 83.3\% \\
        \footnotesize{SparseVLM} & 52.7 & 1505 & 51.8 & 62.2 & 68.2 & 75.1 & 86.5\% & 50.6 & 1402 & 22.7 & 69.0 & - & 65.0 & 72.9\% \\
        \footnotesize{PDrop} & 47.5 & 1561 & 50.6 & 69.0 & 69.2 & 55.9 & 83.3\% & 54.1 & 1247 & 55.3 & 73.1 & 70.8 & 66.1 & 85.0\% \\
        \footnotesize{VisionZip} & 55.1 & \underline{1690} & \underline{55.5} & 69.0 & 72.4 & 77.0 & 92.7\% & 56.2 & 1676 & \underline{57.4} & \textbf{74.4} & 73.7 & 76.0 & 92.9\% \\
        \footnotesize{VisionZip\ddag} & \underline{57.0} & \textbf{1756} & \textbf{56.0} & 68.8 & 74.2 & \underline{80.9} & 95.1\% & 58.1 & 1671 & \textbf{58.5} & 72.3 & 75.2 & 81.6 & 94.6\% \\
        \hline
        \textbf{\footnotesize{CROP-ILP}} & \underline{59.6} & 1675 & 54.9 & \underline{71.5} & \underline{74.8} & \textbf{83.6} & \textbf{96.0\%} & \underline{60.4} & \textbf{1708} & 56.8 & 73.8 & \textbf{76.0} & \textbf{84.8} & \textbf{96.2\%} \\
        \textbf{\footnotesize{CROP-PLC}} & \textbf{60.3} & 1634 & 55.2 & \textbf{71.8} & \textbf{75.0} & 80.2 & \underline{95.4\%} & \textbf{61.1} & \underline{1693} & 57.2 & \underline{73.9} & \underline{75.7} & \underline{81.8} & \underline{95.8\%} \\
        \hline
    \end{tabular}
    }
    \vspace{-1mm}
    \caption{\textbf{Performance of CROP on LLaVA-1.5 compared to existing methods} under three different pruning ratios. The \textbf{bold numbers} indicate the best performance achieved by each MLLM, and the \uline{underline numbers} are the second best. Specifically, VisionZip\ddag  refers to fine-tuned version of VisionZip, where the projector has been fine-tuned to align the pruned visual tokens with the semantic space of the LLM.}
    \label{table: llava1.5}
    \vspace{-3mm}
\end{table*}

\subsection{Training Details}
We constructed the training dataset based on the datasets from LLaVA \citep{liu2023visualinstructiontuning} and Visual-cot \citep{shao2024visualcotadvancingmultimodal}. The composition of this dataset is detailed in Table~\ref{table: datasets}. All training procedures for CROP were conducted on a server equipped with four NVIDIA GeForce RTX A100 GPUs. The training process consists of two main stages:

\paragraph{Stage 1: Localization Module Training} The Qwen2VL-2B model was fine-tuned to function as the Localization Module. During this stage, the visual encoder of Qwen2VL-2B was kept frozen, while all other components were trained on our Region Localization data and Preserving Instruction Following Ability data. Fine-tuning was performed for 2 epochs using the AdamW optimizer with a learning rate of 2e-5. This stage typically completed in approximately 12 hours.

\paragraph{Stage 2: Pre-LLM Compression Module and VLM Co-Fine-tuning} For the PLC strategy, the LLaVA-1.5 model was co-fine-tuned with the newly introduced PLC components. As in Stage 1, LLaVA's visual encoder remained frozen. The learnable parameters of the PLC module, along with LLaVA's projector and LLM backbone, were jointly optimized. This co-fine-tuning utilized our complete training set. Specifically, Region Localization samples with ground truth bounding box annotations were used to train the compression module and help the LLM adapt to the compressed visual inputs, while the LLaVA Instruction Tuning data was used to preserve the model’s original instruction-following capability. Fine-tuning was conducted for 2 epochs using the AdamW optimizer with a learning rate of 1e-5. This stage typically required approximately 26 hours.

\subsection{Experimental Setup}

We evaluated our CROP method on LLaVA-1.5-7B and LLaVA-1.5-13B, conducting extensive ablation studies to verify the importance of contextual region preservation for model perception. Experiments were performed across multiple benchmarks (see Appendix for dataset and metric details), with all evaluations adhering to official dataset guidelines and LLaVA's metrics.

For the ILP strategy, pruning was implemented at layer K=2 of the target VLM. Based on the critical regions identified by our Localization Module, visual token pruning rates were averaged to 66.7\%, 77.8\%, and 88.9\% across the test set. A crucial aspect of this setup is that the retained tokens consistently form continuous rectangular image regions.

For the PLC strategy, the number of learnable queries for contextual regions $N_{Q^{k}}$ and non-contextual regions $N_{Q^{nk}}$ were configured to 64 and 4, respectively. The number of anchor-preserved tokens $ N_{r} $ from the contextual region was set to 64.

\begin{figure*}[t]
    \centering
    \includegraphics[width=0.98\linewidth]{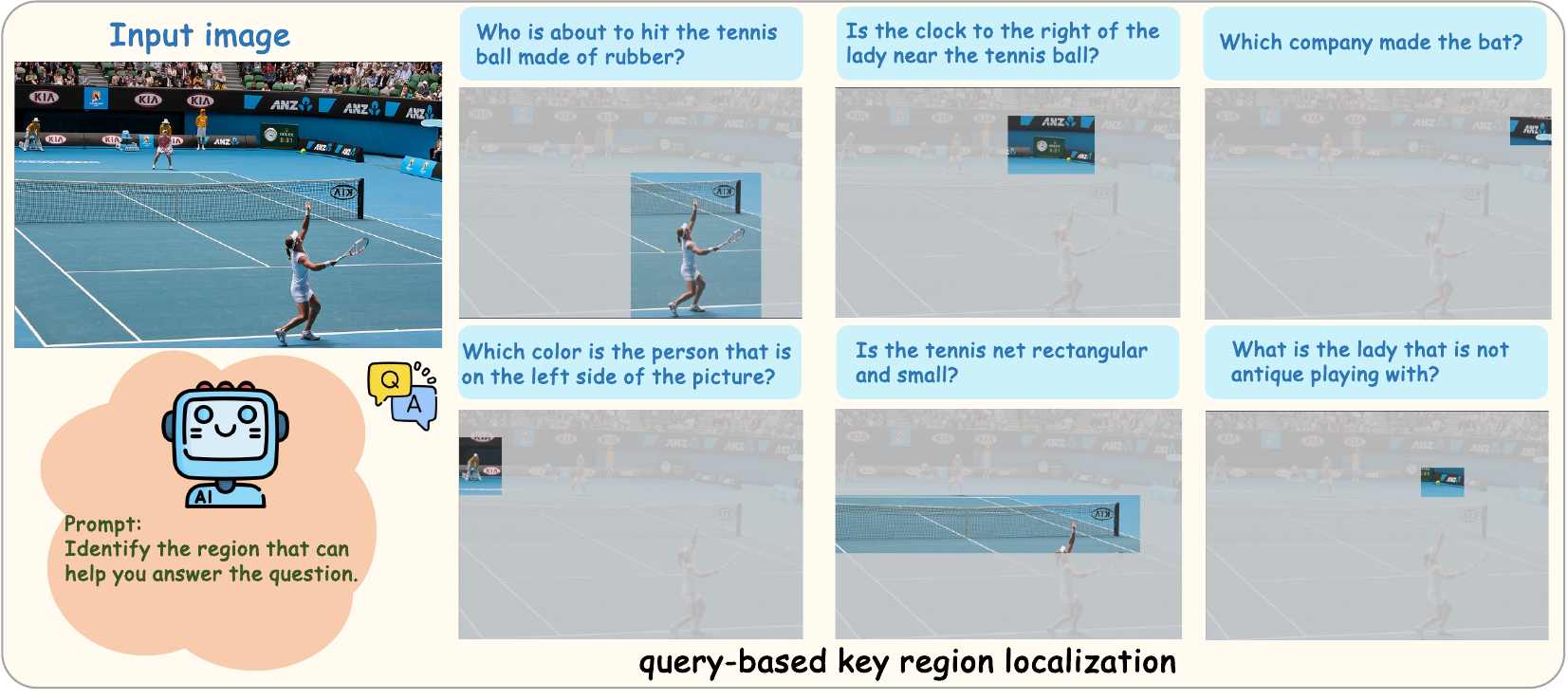}
    \caption{Qualitative examples of query-based contextual region localization by our Localization Module. For different questions, the module identifies specific, contiguous regions (highlighted) crucial for answering, demonstrating focused attention compared to potentially diffuse attention in other methods.}
    \label{figure: Localization}
    \vspace{-5mm}
\end{figure*}

\subsection{Main Results}

\paragraph{Results on LLaVA-1.5.} The primary performance metrics of CROP are presented in Table~\ref{table: llava1.5}. Across the tested pruning ratios, CROP consistently maintained robust visual comprehension capabilities, achieving SOTA results on several benchmarks.

With the ILP strategy, at a 66.7\% visual token pruning rate, LLaVA-1.5-7B and LLaVA-1.5-13B retained 98.9\% and 99.8\% of their respective baseline performances. Even under a more aggressive 88.9\% pruning rate, these models maintained 96.0\% and 96.2\% of their original efficacy. Notably, these ILP results were achieved without any fine-tuning of the LLaVA models.

Our PLC strategy, evaluated on LLaVA-1.5, demonstrated 95.4\% and 95.8\% performance retention relative to its baseline at an 88.9\% visual token pruning rate. These findings strongly validate the effectiveness of CROP, underscoring the benefits of preserving contextual regional information in visual token compression.

\paragraph{Results on LLaVA-NeXT.} To further validate the effectiveness of the proposed CROP framework, we conduct experiments on the LLaVA-NeXT series of models. Unlike LLaVA-1.5, LLaVA-NeXT partitions each input image into four sub-images, which are then processed alongside the original image by the visual encoder—effectively increasing the input to five images per instance. While this design enhances visual perception, it also introduces substantial redundancy and computational overhead. To improve inference efficiency, we apply contextual region-based pruning to the visual tokens encoded from both the original image and its sub-images. We evaluate CROP under three different token retention settings to systematically assess its performance advantages. As shown in Table~\ref{table: llavanext-7b}, our proposed CROP framework consistently maintains strong performances.
\begin{table}[t]
    \begin{center}
        \setlength{\tabcolsep}{1.5pt}
        \renewcommand{\arraystretch}{1.2}
        \footnotesize
        \centering
        \begin{tabular}{l|ccccc|c}
            \shline
            \textbf{Method}\rule{0pt}{2.3ex} & 
            \textbf{GQA} & \textbf{MME} & \textbf{VQA}$^{\text{T}}$ & \textbf{SQA} & \textbf{VQA}$^{\text{V2}}$ &  \textbf{RelAcc} \\
            \shline
            \rowcolor{mygray}
            \multicolumn{7}{c}{\footnotesize{\textit{Upper Bound, 576 Tokens} \ (\textbf{100\%})}} \\
            LLaVA-NeXT & 64.2 & 1842 & 61.3 & 70.2 & 80.1 & 100\% \\
            \rowcolor{mygray}
            \multicolumn{7}{c}{\textit{Retain Averaged 640 Tokens \ $\fg{(\downarrow 77.8\%)}$}} \\
            FastV & 62.0 & \textbf{1807} & 60 & 69.1 & 79.5 & 98.0\% \\
            SparseVLM & 60.3 & 1772 & 57.8 & 67.7 & 77.1 & 95.4\% \\
            VisionZip & 61.3 & 1787 & 60.2 & 68.1 & 79.1 & 97.3\% \\
            VisionZip\ddag & 62.4 & 1778 & \textbf{60.8} & 67.9 & 79.9 & 97.9\% \\
            \hline
            \textbf{CROP-ILP} &  \textbf{63.2} &  1768 &  60.0 &  \textbf{69.6} &  \textbf{80.0} &  \textbf{98.3\%} \\
            \hline
            \rowcolor{mygray}
            \multicolumn{7}{c}{\textit{Retain Averaged 320 Tokens \ $\fg{(\downarrow 88.9\%)}$}} \\
            FastV & 54.9 & 1539 & 54.8 & 68.2 & 69.6 & 88.5\% \\
            SparseVLM & 57.7 & 1694 & 55.9 & 67.3 & 73.4 & 92.1\% \\
            VisionZip & 59.3 & 1702 & 58.9 & 67.3 & 76.2 & 94.4\% \\
            VisionZip\ddag & 61.0 & \textbf{1770} & \textbf{59.3} & 67.5 & 78.4 & 96.4\% \\
            \hline
            \textbf{CROP-ILP} & \textbf{62.0} & 1749 & 59.1 & \textbf{69.1} & \textbf{78.7} & \textbf{96.9\%} \\
            \hline
            \rowcolor{mygray}
            \multicolumn{7}{c}{\textit{Retain Averaged 160 Tokens \ $\fg{(\downarrow 94.4\%)}$}} \\
            FastV & 49.3 & 1496 & 47.5 & 67.9 & 68.2 & 83.5\% \\
            SparseVLM & 51.2 & 1542 & 46.4 & 67.5 & 66.3 & 83.6\% \\
            VisionZip & 55.5 & 1630 & 56.2 & 68.3 & 71.4 & 90.6\% \\
            VisionZip\ddag & 58.2 & \textbf{1699} & 57.3 & 67.5 & 75.6 & 93.4\% \\
            \hline
            \textbf{CROP-ILP} & \textbf{60.2} & 1668 & \textbf{57.7} & \textbf{68.7} & \textbf{76.1} & \textbf{94.3\%} \\
            \shline
        \end{tabular}
        \vspace{-1mm}
        \caption{Performance of CROP-ILP with LLaVA-NeXT-7B on Various VLM Benchmarks. The \textbf{bold values} indicate the best performance.}
        \label{table: llavanext-7b}
    \end{center}
    \vspace{-3mm}
\end{table}

\subsection{Efficiency Analysis}

Owing to the workflow design of CROP, the Localization Module and the VLM backbone operate relatively independently during inference. While the Localization Module introduces a non-negligible number of parameters, it remains lightweight compared to the VLM backbone and achieves faster inference. Consequently, when handling multiple batches or video streams, the VLM backbone only needs to wait for the Localization Module once, typically during the first batch. After this initial step, the localization module and the VLM backbone can process subsequent inputs in parallel. Thanks to this “ahead-of-time prediction” property, the VLM backbone is freed from serial waiting for localization outputs, thereby reducing overall latency. To validate this claim, we conducted speed benchmarking experiments on the LLaVA-NeXT model. As shown in the Table~\ref{table: efficiency}, our ILP method accelerates LLaVA-NeXT-13B by more than twofold and even surpasses the speed of the smaller LLaVA-NeXT-7B. These results demonstrate the effectiveness of CROP in enhancing efficiency. Additional analyses on efficiency are provided in the Appendix.

\section{Analyses and Discussion}
Previous visual token compression methods have largely overlooked the importance of preserving contextual regions during compression. This section analyzes the impact of our CROP strategy on model perception and understanding, and further evaluates the effectiveness of our Localization Module.

\begin{table}[t]
    \begin{center}
        \setlength{\tabcolsep}{1.5pt}
        \renewcommand{\arraystretch}{1.2}
        \footnotesize
        \centering
        \begin{tabular}{l|ccccc}
            \shline
            \textbf{Method}\rule{0pt}{2.3ex} & 
            \textbf{GQA} & \textbf{MME} & \textbf{VQA}$^{\text{T}}$ & \textbf{SQA} & \textbf{VQA}$^{\text{V2}}$ \\
            \shline
            \rowcolor{mygray}
            \multicolumn{6}{c}{\footnotesize{\textit{Upper Bound, 576 Tokens} \ (\textbf{100\%})}} \\
            LLaVA-NeXT-7B & 3356s & 606s & 1611s & 874s & 2336s \\
            LLaVA-NeXT-13B & 5394s & 953s & 2497s & 1349s & 3671s \\
            \rowcolor{mygray}
            \multicolumn{6}{c}{\textit{Retain Averaged 640 Tokens \ $\fg{(\downarrow 77.8\%)}$}} \\
            FastV & 2857s & 516s & 1401s & 860s & 1928s \\
            CROP-ILP & 2833s & 497s & 1400s & 850s & 1861s \\
            \rowcolor{mygray}
            \multicolumn{6}{c}{\textit{Retain Averaged 160 Tokens \ $\fg{(\downarrow 94.4\%)}$}} \\
            FastV & 1973s & 350s & 1029s & 580s & 1303s \\
            CROP-ILP & 2028s & 355s & 1066s & 592s & 1330s \\
            \shline
        \end{tabular}
        \vspace{-1mm}
        \caption{Efficiency analysis of CROP-ILP on LLaVA-NeXT-13B. The table reports the total inference time on multiple VQA benchmarks, comparing the baseline LLaVA-NeXT-7B/13B with FastV and our CROP-ILP applied to LLaVA-NeXT-13B.}
        \label{table: efficiency}
    \end{center}
    \vspace{-3mm}
\end{table}

\begin{table}[t]
    \begin{center}
        \setlength{\tabcolsep}{2.3pt}
        \renewcommand{\arraystretch}{1.2}
        \footnotesize
        \centering
        \begin{tabular}{l|ccccc|c}
            \shline
            \textbf{Method}\rule{0pt}{2.3ex} & 
            \textbf{GQA} & \textbf{MME} & \textbf{VQA}$^{\text{T}}$ & \textbf{SQA} & \textbf{POPE} &  \textbf{RelAcc} \\
            \shline
            \rowcolor{mygray}
            \multicolumn{7}{c}{\textit{Upper Bound, 576 Tokens} \ $\textbf{(100\%)}$} \\
            LLaVA-1.5 & 61.9 & 1862 & 58.2 & 69.5 & 85.9 & 100\%  \\
            \rowcolor{mygray}
            \multicolumn{7}{c}{\footnotesize{\textit{Retain Averaged 192 Tokens} \ $\fg{(\downarrow 66.7\%)}$}} \\
            prune & 58.7 & 1735 & 56.8 & \textbf{69.2} & \textbf{85.4} & 96.9\% \\ 
			\textbf{CROP-ILP} &  \textbf{61.3} & \textbf{1817} & \textbf{57.4} & \textbf{69.2} & 85.3 & \textbf{98.8\%}\\ 
            \rowcolor{mygray}
            \multicolumn{7}{c}{\footnotesize{\textit{Retain Averaged 128 Tokens} \ $\fg{(\downarrow 77.8\%)}$}} \\
            prune & 57.9 & 1734 & 56.2 & \textbf{69.6} & \textbf{84.6} & 96.4\% \\ 
			\textbf{CROP-ILP} &  \textbf{60.8} & \textbf{1771} & \textbf{56.9} & 69.5 & 84.4 & \textbf{97.9\%}\\
            \rowcolor{mygray}
            \multicolumn{7}{c}{\footnotesize{\textit{Retain Averaged 64 Tokens} \ $\fg{(\downarrow 88.9\%)}$}} \\
            prune & 55.0 & 1625 & 54.1 & 69.9 & \textbf{83.8} & 93.4\% \\ 
			\textbf{CROP-ILP} &  \textbf{59.6} & \textbf{1675} & \textbf{54.9} & \textbf{71.5} & 83.6 & \textbf{96.2\%}\\
            \shline
        \end{tabular}
        \vspace{-1mm}
       \caption{Performance of LLaVA-1.5-7B with "prune" Guided by the Localization Module. Results are shown across various VQA benchmarks and pruning rates, compared to baseline LLaVA-1.5-7B and representative discrete token pruning methods. The \textbf{bold values} indicate the best performance.}
	\label{table: ablation_localization}
    \end{center}
    \vspace{-3mm}
\end{table}

\begin{table}[t]
    \begin{center}
        \setlength{\tabcolsep}{2.3pt}
        \renewcommand{\arraystretch}{1.2}
        \footnotesize
        \centering
        \begin{tabular}{l|ccccc|c}
            \shline
            \textbf{Method}\rule{0pt}{2.3ex} & 
            \textbf{GQA} & \textbf{MME} & \textbf{VQA}$^{\text{T}}$ & \textbf{SQA} & \textbf{POPE} &  \textbf{RelAcc} \\
            \shline
            \rowcolor{mygray}
            \multicolumn{7}{c}{\textit{Upper Bound, 576 Tokens} \ $\textbf{(100\%)}$} \\
            LLaVA-1.5 & 61.9 & 1862 & 58.2 & 69.5 & 85.9 & 100\%  \\
            \rowcolor{mygray}
            \multicolumn{7}{c}{\textit{Retain Averaged 64 Tokens} \ $\fg{(\downarrow 88.9\%)}$}\\
			\textbf{w/o $X^{r}$} &  56.4 & 1593 & 52.3 & 67.6 & 81.2 & 91.7\%\\ 
			\textbf{CROP-PLC} &  \textbf{60.3} & \textbf{1634} & \textbf{56.2} & \textbf{71.8} & \textbf{83.7} & \textbf{96.5\%}\\
            \shline
        \end{tabular}
        \vspace{-1mm}
       \caption{Ablation Study on the Impact of Anchor-Preserved Tokens $X^{r}$ in the PLC Strategy on LLaVA-1.5-7B. Performance is compared between the full CROP-PLC and CROP-PLC(w/o $X^{r}$). The results demonstrate that when the anchor tokens $X^{r}$ are discarded, the model's performance declines as it then processes visual tokens from the key region that are effectively more fragmented. The \textbf{bold values} indicate the best performance.}
	\label{table: ablation_region}
    \end{center}
    \vspace{-3mm}
\end{table}

\subsection{Localization Module's Token Selection}

To validate the efficacy of our Localization Module, we conducted experiments on LLaVA-1.5-7B. Using the contextual regions identified by the module, we derived the corresponding visual token indices. Tokens outside these contextual regions were pruned entirely before being fed to the LLM, an approach we denote as \textbf{prune}. We tested this at visual token pruning rates of 66.7\%, 77.8\%, and 88.9\%.

As shown in Table~\ref{table: ablation_localization}, even when relying solely on visual information from these localized contextual regions, the model maintained robust performance. Notably, this simple region-only approach surpassed several meticulously designed discrete token pruning strategies. These results indicate two key findings: (1) our Localization Module accurately identifies critical, query-relevant regions across diverse benchmarks, and (2) retaining tokens from these contextual regions preserves the majority of essential visual information more effectively than discrete token selection methods.

\subsection{Visual Token Selection Strategies}

Discrete token pruning strategies, especially those reliant on cross-modal attention, can suffer from attention diffusion. This may lead to the retention of spatially disjointed or peripherally relevant visual tokens that offer limited effective visual guidance. In contrast, CROP’s approach, by identifying and preserving contextual regions as illustrated in Figure~\ref{figure: Localization}, inherently leverages the contiguity of visual objects and query-specific context. This method focuses the selection on principal visual elements relevant to the query, maintaining spatial coherence and naturally filtering out irrelevant background or edge information. Consequently, CROP yields a more compact and semantically rich set of visual tokens, leading to the improved performance retention validated in our main results.

\subsection{Impact of Contextual Region Preservation on Perception}

To further quantify the importance of explicitly preserving contextual regional information within our PLC framework, we conducted an ablation study on LLaVA-1.5-7B. We trained a variant of the PLC architecture that omits the anchor-preserved tokens $X^{r} $ and the associated fusion step (Equation~\ref{eq:plc_fuse_attn}). In this ablated setup, the visual input to the LLM comprised only the concatenated compressed representations from contextual regions $\hat{X}^{kv}$ and non-contextual regions $\hat{X}^{nkv}$.

As detailed in Table~\ref{table: ablation_region}, removing the anchor-preserved tokens $X^{r} $ led to a notable performance decrease of approximately 5\% on average across benchmarks when compared to the full PLC strategy. Furthermore, this ablated PLC variant sometimes underperformed the simpler "prune" strategy (Section 5.1). This finding strongly supports our assertion that explicitly preserving a core set of coherent regional tokens, via mechanisms like the $X^{r} $ fusion in PLC, is crucial for maintaining the model's perceptual capabilities, especially under aggressive token compression. This finding suggests that for VLMs, preserving the structural integrity of key visual information is as critical as retaining individual, semantically important tokens. This insight—that spatial coherence matters—is a key consideration that could inform the design of future token reduction techniques.

\section{Conclusion}

This work addresses the computational inefficiency in VLMs caused by excessive visual tokens. We introduced CROP, a framework that dramatically reduces token count by first identifying and then preserving contiguous, query-relevant visual regions. Our experiments demonstrate that CROP significantly outperforms existing methods. Notably, our training-free ILP strategy achieves competitive performance, offering an efficient drop-in solution for existing VLMs. Our findings underscore a critical principle for future VLM design: preserving the spatial integrity of key visual information is essential for building efficient and perceptually robust models.

\section*{Limitation}

Despite the remarkable efficiency demonstrated by CROP, particularly with its training-free ILP strategy, the computational gains are primarily observed during inference. The initial fine-tuning of the PLC module, while a one-time cost, does add to the overall model development pipeline. Future work might explore methods to further streamline this initial training phase or develop entirely training-free compression strategies, both internally and externally, such as meta-learning universal compression approaches that require minimal to no task-specific fine-tuning, thereby achieving even more efficient and true "zero-shot" compression.

\section*{Acknowledgements}
We would like to express our gratitude to the anonymous reviewers for their insightful feedback and constructive suggestions. This research was supported by grants from the National Natural Science Foundation of China (No. 62476275).

\bibliography{custom}

\clearpage

\appendix
\section*{Appendix}

\section{Template of the Training Data Examples}\label{appendix:data}

We now describe our training data format. Both the Localization Module and the PLC strategy's compression module are trained using examples where contextual regions are defined by the 8x8 grid-based block coordinates presented in Section 3.2. Table~\ref{table: datasets} provides further details on the datasets used.

\begin{tcolorbox}[colback=lightgray!50!white,colframe=lightgray,title=\textcolor{black}{Template of Training Example for Region Selection Token}, width=\columnwidth]
\footnotesize
\textbf{User}: $<$image$>$ $<$question$>$ Please identify the region that can help you answer the question better.

\noindent\textbf{Assistant}: $<$x\_min$>$ $<$y\_min$>$ $<$x\_max$>$ $<$y\_max$>$.
\end{tcolorbox}

\section{Details of Evaluation Benchmarks}\label{appendix:benchmarks}

In this section, we provide a brief overview of the benchmarks used in our experiments.

\vspace{3pt}
\noindent\textbf{GQA} \cite{hudson2019gqanewdatasetrealworld} serves as a benchmark for evaluating visual scene understanding and reasoning capabilities. It utilizes a combination of images, associated questions, and scene graphs, specifically testing a model's ability to grasp spatial relationships and object properties within intricate visual contexts.

\vspace{3pt}
\noindent\textbf{MME} \cite{fu2024mmecomprehensiveevaluationbenchmark} offers a comprehensive assessment of model performance across 14 distinct subtasks, which investigate both perceptual abilities and cognitive skills. The benchmark employs meticulously designed instruction-answer pairs to ensure an equitable and thorough evaluation of a model's multimodal competence. The final reported score represents the aggregate of the perception and cognition scores.

\vspace{3pt}
\noindent\textbf{TextVQA} \cite{DBLP:conf/cvpr/SinghNSJCBPR19} is designed to gauge a model's capacity to interpret and reason about textual elements found within images. By demanding the synthesis of visual and textual data, it stands as an important benchmark for assessing text-oriented reasoning in visual environments. For conciseness in our experimental tables, we refer to it as ``VQA$^{\text{T}}$''.

\vspace{3pt}
\noindent\textbf{ScienceQA} \cite{lu2022learnexplainmultimodalreasoning} encompasses a broad range of scientific disciplines, including natural, language, and social sciences. Its questions are structured into 26 topics, 127 categories, and 379 skills. This benchmark evaluates a model's multimodal comprehension, aptitude for multi-step reasoning, and its interpretability, thus offering a robust framework for assessing the application of scientific knowledge in visual contexts. In our experiments, we focus exclusively on the image-based samples, denoted as ``SQA$^{\text{I}}$'' in the experimental tables.

\vspace{3pt}
\noindent\textbf{VQA-v2} \cite{goyal2017makingvvqamatter} is a comprehensive benchmark consisting of 265,000 images that capture real-world scenes and objects. Each image is presented with open-ended questions, and for each question, ten ground truth answers provided by humans are available.

\vspace{3pt}
\noindent\textbf{POPE} \cite{yue2024mmmumassivemultidisciplinemultimodal} is focused on evaluating object hallucination by presenting binary questions concerning the existence of objects in images. It utilizes metrics such as Accuracy, Recall, Precision, and F1 score, applied over three different sampling methodologies. The score reported reflects the mean accuracy across these three methods: adversarial, random, and popular.

\section{Benchmarks and Metrics for Evaluation}\label{appendix:metrics}

The benchmarks utilized in our study, along with their respective evaluation metrics, adhere to the official guidelines provided by each benchmark and the official evaluation scripts from the LLaVA model. For the VQA datasets, we assessed the zero-shot question-answering performance using a single input image. All results reported in this paper are averaged across multiple experimental runs.

\section{More Experimental Results}\label{appendix:results}
\subsection{Results on Qwen}

After demonstrating the effectiveness of the CROP framework on the LLaVA family of models, we further evaluate its performance on the Qwen series. The Qwen2-VL model adaptively determines the number of visual tokens based on the resolution of the input image, allowing it to process high-resolution inputs and exhibit strong visual understanding capabilities. As shown in Table~\ref{table: qwen2vl-7b}, we evaluate the performance of CROP-ILP under four different token retention settings. Even with pruning rates ranging from 50\% to 77.8\%, Qwen2-VL-7B retains between 98.1\% and 95.8\% of the original performance. These results demonstrate the efficiency and robustness of our method on the Qwen architecture, and further highlight the strong generalization ability of the proposed CROP framework across diverse multimodal model families.
\begin{table}[t]
    \begin{center}
        \setlength{\tabcolsep}{2.3pt}
        \renewcommand{\arraystretch}{1.2}
        \footnotesize
        \centering
        \begin{tabular}{l|ccccc|c}
            \shline
            \textbf{Method}\rule{0pt}{2.3ex} & 
            \textbf{GQA} & \textbf{MME} & \textbf{VQA}$^{\text{T}}$ & \textbf{SQA} & \textbf{POPE} &  \textbf{RelAcc} \\
            \shline
            \rowcolor{mygray}
            \multicolumn{7}{c}{\footnotesize{\textit{Dynamic Number of Tokens} \ (\textbf{100\%})}} \\
            Qwen2-VL & 61.58 & 2269 & 83.92 & 84.34 & 88.35 & 100\% \\
            \rowcolor{mygray}
            \multicolumn{7}{c}{\textit{Retain Averaged 50\% Tokens}} \\
            FastV & 55.50 & \textbf{2251} & 79.90 & 78.90 & 82.99 & 94.4\% \\
            \textbf{CROP-ILP} & \textbf{61.37} & 2143 & \textbf{81.80} & \textbf{83.42} & \textbf{88.41} & \textbf{98.1\%} \\
            \rowcolor{mygray}
            \multicolumn{7}{c}{\textit{Retain Averaged 40\% Tokens}} \\
            FastV & 55.58 & \textbf{2216} & 76.52 & 77.46 & 81.16 & 92.6\% \\
            \textbf{CROP-ILP} & \textbf{60.67} & 2066 & \textbf{81.61} & \textbf{83.07} & \textbf{88.23} & \textbf{97.0\%} \\
            \rowcolor{mygray}
            \multicolumn{7}{c}{\textit{Retain Averaged 33.3\% Tokens}} \\
            FastV & 54.64 & \textbf{2174} & 73.12 & 75.13 & 80.36 & 90.3\% \\
            \textbf{CROP-ILP} & \textbf{60.21} & 2050 & \textbf{80.61} & \textbf{82.62} & \textbf{88.06} & \textbf{96.4\%} \\
            \rowcolor{mygray}
            \multicolumn{7}{c}{\textit{Retain Averaged 22.2\% Tokens}} \\
            FastV & 53.30 & \textbf{2041} & 71.56 & 72.84 & 76.15 & 86.9\% \\
            \textbf{CROP-ILP} & \textbf{59.85} & 2013 & \textbf{80.34} & \textbf{82.60} & \textbf{87.86} & \textbf{95.8\%} \\
            \shline
        \end{tabular}
        \vspace{-1mm}
        \caption{Performance of CROP-ILP with Qwen2-VL-7B on Various VLM Benchmarks. The \textbf{bold values} indicate the best performance.}
		\label{table: qwen2vl-7b}
    \end{center}
    \vspace{-3mm}
\end{table}

\subsection{Ablation Study on the Pruning Layer}

In the aforementioned CROP-ILP experiments, we set the pruning layer K=2 to ensure a fair comparison with other methods that prune visual tokens in the early layers of the model. To further examine the impact of pruning at different layers, we conducted an ablation study on LLaVA-1.5-7B using the ILP method, where the visual tokens were pruned to an average of 64. As shown in the Table~\ref{table: ablation_K}, increasing K allows the VLM to retain more visual information, resulting in a general improvement in overall performance, and in fact, pruning at the top layers can even lead to performance gains. On the other hand, pruning in the early layers yields substantial efficiency improvements but inevitably introduces minor performance drops. In practical scenarios, one may balance these trade-offs and select an appropriate pruning layer K according to the application requirements.

\begin{table}[t]
    \begin{center}
        \setlength{\tabcolsep}{0.8pt}
        \renewcommand{\arraystretch}{1.2}
        \footnotesize
        \centering
        \begin{tabular}{l|ccccc|c}
            \shline
            \textbf{Method}\rule{0pt}{2.3ex} & 
            \textbf{GQA} & \textbf{MME} & \textbf{VQA}$^{\text{T}}$ & \textbf{SQA} & \textbf{POPE} &  \textbf{RelAcc} \\
            \shline
            \rowcolor{mygray}
            \multicolumn{7}{c}{\textit{Upper Bound, 576 Tokens} \ $\textbf{(100\%)}$} \\
            LLaVA-1.5 & 61.94 & 1862 & 58.20 & 69.54 & 85.93 & 100\%  \\
            \rowcolor{mygray}
            \multicolumn{7}{c}{\textit{Retain Averaged 64 Tokens} \ $\fg{(\downarrow 88.9\%)}$}\\
		    CROP-ILP(K=32) &  61.96 & 1866 & 58.18 & 70.95 & 85.93 & 100.5\%\\
		    CROP-ILP(K=31) &  61.94 & 1869 & 57.92 & 70.95 & 85.99 & 100.4\%\\
		    CROP-ILP(K=30) &  61.99 & 1855 & 58.04 & 70.97 & 86.01 & 100.3\%\\
		    CROP-ILP(K=25) &  61.97 & 1854 & 57.77 & 70.97 & 86.09 & 100.2\%\\
		    CROP-ILP(K=20) &  61.63 & 1871 & 57.48 & 70.93 & 86.09 & 100.2\%\\
		    CROP-ILP(K=15) &  61.35 & 1869 & 56.60 & 71.00 & 85.84 & 99.7\%\\
		    CROP-ILP(K=10) &  60.11 & 1749 & 55.78 & 70.88 & 83.68 & 97.2\%\\
		    CROP-ILP(K=7) &  59.92 & 1775 & 54.80 & 71.47 & 84.28 & 97.4\%\\
		    CROP-ILP(K=5) &  59.62 & 1713 & 54.59 & 71.47 & 83.71 & 96.5\%\\
		    CROP-ILP(K=3) &  59.58 & 1670 & 54.46 & 71.23 & 83.69 & 95.9\%\\
		    CROP-ILP(K=2) &  59.59 & 1675 & 54.92 & 71.54 & 83.57 & 96.1\%\\ 
		    CROP-ILP(K=1) &  59.62 & 1665 & 54.24 & 70.88 & 83.55 & 95.6\%\\
		    prune &  55.03 & 1625 & 54.13 & 69.92 & 83.84 & 93.5\%\\
            \shline
        \end{tabular}
        \vspace{-1mm}
       \caption{Ablation study on the pruning layer K in the LLaVA-1.5-7B model. The results demonstrate that applying the CROP-ILP method for pruning across different layers of the model largely preserves its original performance.}
	\label{table: ablation_K}
    \end{center}
    \vspace{-3mm}
\end{table}

\subsection{Robustness of the Localization Module}
To validate the robustness of our Localization Module, we randomly selected 3000 samples from TextVQA set and manually annotated the ground truth regions. In the following, $GT$ denotes the manually labeled ground truth region, and $Pred$ denotes the region predicted by our Localization Module. We used Area-based Recall to quantify localization quality, which is defined as the ratio of the intersection area between $GT$ and $Pred$ to the area of $GT$. This metric reflects how accurately the predicted region capture the true relevant visual context. As shown in the Table~\ref{table: recall}, the results show that our Localization Module achieves high perception accuracy, with strong overlap between the predicted and human-annotated contextual regions.

\begin{table}[t]
    \centering
    \resizebox{1\linewidth}{!}{%
    \begin{tabular}{@{}l l l l l@{}}
    \toprule
    \textbf{Sample Count} & \textbf{Mean Recall} & \textbf{Recall \textgreater 0.5} & \textbf{Recall \textgreater 0.7} & \textbf{Recall \textgreater 0.9}\\ 
    \midrule
    3000 & 0.9477 & 2871 & 2802 & 2706\\
    \bottomrule
    \end{tabular}%
    }
    \caption{The Area-based Recall value of the contextual region localized by the Localization Module.}
    \label{table: recall}
\end{table}
\begin{table}[t]
    \centering
    \resizebox{1\linewidth}{!}{%
    \renewcommand{\arraystretch}{1.5} 
    \begin{tabular}{@{}l l l l l@{}}
    \hline
    \textbf{LLaVA-1.5-7B baseline} & $\boldsymbol{GT}$ & $\boldsymbol{Pred}$ & $\boldsymbol{Center}$ & $\boldsymbol{Random}$\\ 
    \hline
    52.32 & 53.20 & 53.14 & 47.35 & 45.76\\
    \hline
    \textbf{LLaVA-1.5-13B baseline} & $\boldsymbol{GT}$ & $\boldsymbol{Pred}$ & $\boldsymbol{Center}$ & $\boldsymbol{Random}$\\ 
    \hline
    55.90 & 55.85 & 55.81 & 49.78 & 48.47\\
    \hline
    \textbf{Qwen2VL-7B baseline} & $\boldsymbol{GT}$ & $\boldsymbol{Pred}$ & $\boldsymbol{Center}$ & $\boldsymbol{Random}$\\ 
    \hline
    83.75 & 81.93 & 81.72 & 66.44 & 62.67\\
    \hline
    \end{tabular}%
    }
    \caption{The performance of the CROP-ILP strategy on LLaVA and Qwen models with different region selection strategies.}
    \label{table: localization_quality}
\end{table}

To measure how localization quality affects model performance, we selected two alternative regions in each image: a center region($Center$) and a randomly chosen region($Random$), both resized to the same size as the predicted region. We then applied ILP pruning with pruning layer $K$ = 2 on several VLMs. The results are shown in the Table~\ref{table: localization_quality}.

They demonstrate that localization quality has a substantial impact on pruning performance. The predicted regions from our Localization Module closely match the performance of $GT$ regions, whereas the $Center$ and $Random$ regions cause significant drops in model performance.

\section{More Visualized Results}\label{appendix:visualized}

\begin{figure*}[t]
    \centering
    \includegraphics[width=0.98\linewidth]{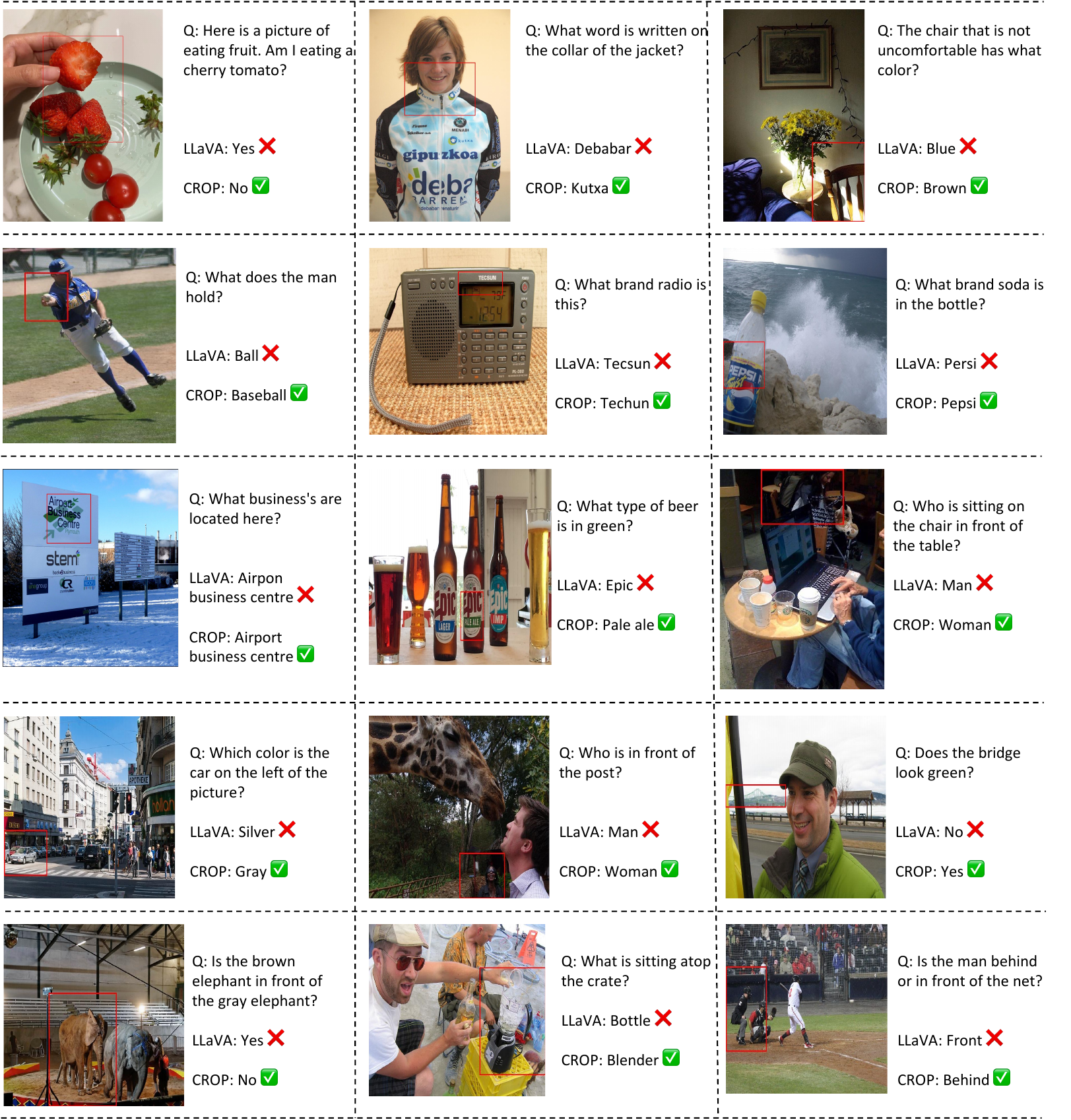}
    \caption{Examples of the CROP-ILP strategy applied to the LLaVA-1.5-7B model, including some routine questions and questions involving object relationships. The responses of both the base LLaVA model and the LLaVA model using CROP are presented, with check marks and cross marks indicating whether each response is correct or incorrect.}
    \label{figure: Visualized_Results_a}
    \vspace{-5mm}
\end{figure*}

In this section, we present several visual examples of the CROP method to provide a more intuitive understanding of the importance of contextual regions in visual token pruning. All experiments are conducted using LLaVA-1.5-7B as the base Visual Language Model. We have gathered a collection of both successful and unsuccessful cases, including those involving routine questions, questions involving object relationships, and more global questions. Overall, in the majority of cases, CROP and the LLaVA model demonstrated consistent performance. In tasks where contextual regions are crucial, we found that by helping the VLM locate the key information, it was able to correctly answer questions that would otherwise be incorrectly answered, even after pruning most of the visual tokens. This is because we effectively eliminated unnecessary visual distractions, allowing the VLM to focus on the key visual entities, as shown in Figure~\ref{figure: Visualized_Results_a}.

\begin{figure*}[t]
    \centering
    
    \includegraphics[width=0.98\linewidth]{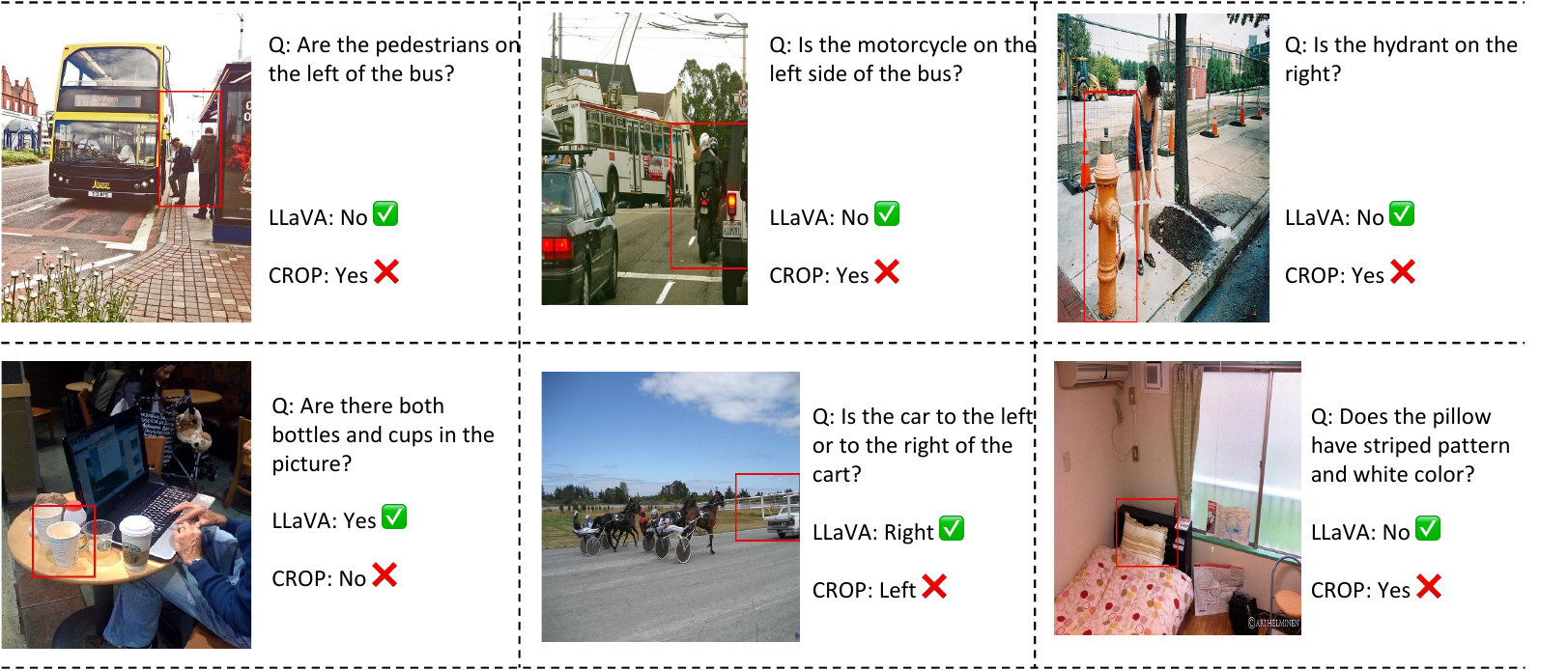}
    \caption{Examples of the CROP-ILP strategy applied to the LLaVA-1.5-7B model, including some questions involving object relationships. The responses of both the base LLaVA model and the LLaVA model using CROP are presented, with check marks and cross marks indicating whether each response is correct or incorrect.}
    \label{figure: Visualized_Results_b}
    \vspace{-5mm}
\end{figure*}

However, for tasks involving object relationships, the VLM requires awareness of multiple objects or parts of them to provide correct answers. If too much visual information is discarded, it impairs the VLM's judgment. We present both successful and unsuccessful cases in this regard. In the three examples at the bottom of Figure~\ref{figure: Visualized_Results_a}, the questions involve two objects. Our Localization Module preserved the most critical information in the contextual region, retaining only part of the objects involved in the relationships. Despite this, the VLM was still able to make the correct judgment and answer accurately. In contrast, in the example shown in Figure~\ref{figure: Visualized_Results_b}, the VLM lost too much object information, preventing it from making an accurate judgment and resulting in an incorrect answer.

\begin{figure*}[t]
    \centering
    \includegraphics[width=0.98\linewidth]{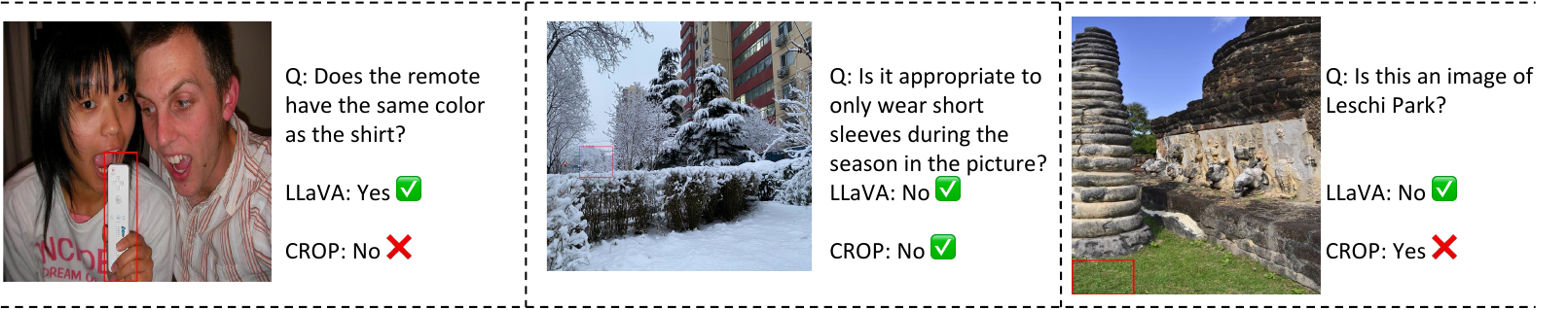}
    \caption{Examples of the CROP-ILP strategy applied to the LLaVA-1.5-7B model, including some questions requiring global information. The responses of both the base LLaVA model and the LLaVA model using CROP are presented, with check marks and cross marks indicating whether each response is correct or incorrect.}
    \label{figure: Visualized_Results_c}
    \vspace{-5mm}
\end{figure*}

Furthermore, for tasks requiring global information, the VLM is likely to provide an incorrect answer if the contextual region does not encompass the necessary global context, as shown in Figure~\ref{figure: Visualized_Results_c}. To address these types of problems, we will explore the use of multiple contextual regions in future work, which will enhance the granularity of localization and help the VLM retain key visual objects as well as global visual information.

\end{document}